\documentclass{bmvc2k}

\usepackage{times}
\usepackage[table,xcdraw]{xcolor}
\usepackage{xcolor}
\usepackage{soul}
\usepackage[utf8]{inputenc}

\usepackage{times}
\usepackage{epsfig}
\usepackage{graphicx}
\usepackage{amsmath}
\usepackage{amssymb}
\usepackage[misc]{ifsym}

\usepackage{float,tabularx}

\usepackage{multirow}
\usepackage{array}
\newcolumntype{L}[1]{>{\raggedright\let\newline\\\arraybackslash\hspace{0pt}}m{#1}}
\newcolumntype{C}[1]{>{\centering\let\newline\\\arraybackslash\hspace{0pt}}m{#1}}
\newcolumntype{R}[1]{>{\raggedleft\let\newline\\\arraybackslash\hspace{0pt}}m{#1}}

\title{SampleAhead: Online Classifier-Sampler Communication for {\em Learning from Synthesized Data}}

\addauthor{Qi Chen, Weichao Qiu, Yi Zhang}{{qchen42,wqiu7,yzh}@jhu.edu}{1}
\addauthor{Lingxi Xie\textsuperscript{(\Letter)}, Alan L. Yuille}{198808xc@gmail.com,alan.yuille@jhu.edu}{1}

\addinstitution{
Department of Computer Science,\\
The Johns Hopkins Univerisity\\
{\scriptsize * This work is supported by IARPA via DOI/IBC contract \#D17PC00345 and ONR grant N00014-15-1-2356.}
}

\runninghead{Chen et.al}{SampleAhead -- Online Classifier-Sampler Communication}

\begin{document}

\maketitle

\begin{abstract}
State-of-the-art techniques of computer vision are mostly data-driven, but collecting and manually labeling a large scale dataset is both difficult and expensive. A promising alternative is to use synthesized training data, so that the dataset size can be significantly enlarged with little human labor. But, this raises an important problem: given an {\bf infinite} data space, how to effectively sample a {\bf finite} subset to train a visual recognition model?

This paper presents an approach for learning from synthesized data effectively. The motivation is straightforward -- increasing the probability of seeing difficult training data. We introduce a module named {\bf SampleAhead} to formulate the learning process into an online communication between a {\em classifier} and a {\em sampler}, and update them iteratively. In each round, we adjust the sampling distribution according to the classification results, and train the classifier using the data sampled from the updated distribution. Experiments are performed by introducing synthesized images rendered from ShapeNet models to assist PASCAL3D+ classification. Our approach enjoys higher classification accuracy, especially in the scenario of a limited number of training samples. This demonstrates its efficiency in exploring the infinite data space.
\end{abstract}

\section{Introduction}
\label{sec:introduction}

Recent progress in computer vision has been boosted by deep neural networks trained with a large amount of labeled data. Researchers made every effort to increase the volume~\cite{deng2009imagenet,wu20143d} and representativeness~\cite{everingham2010pascal} of these datasets, however, the collection and annotation remain labor-intensive and error-prone. A smart idea to address this problem is to generate synthesized data ({\em e.g.}, from a virtual world~\cite{butler2012naturalistic,su2015render}) with a minimal amount of human labor.

But, because the synthesized environment allows us to sample an {\bf infinite} amount of data, an important yet unstudied problem is raised: given a constrained time, how to effectively sample a {\bf finite} subset so as to maximize the performance of a vision system? We address this problem with object pose estimation, a fundamental task in 3D computer vision. Note that for some specific tasks such as object pose estimation, integrating synthesized data helps a lot in improving recognition accuracy, but previous approaches often sampled data uniformly from the synthesized space~\cite{su2015render}, leading to a redundant set of {\em easy} training cases, while the {\em hard} cases cannot get trained sufficiently.

Inspired by previous work~\cite{shrivastava2016training} which adjusted data weights according to their difficulties in an online manner, we suggest a learning system which is composed of two components, with a {\em classifier} (parameterized by a set of network weights $\boldsymbol{\theta}$) dealing with the recognition task, and a {\em sampler} (parameterized by a class distribution $P\!\left(\cdot\right)$ over viewpoint parameters, {\em e.g.}, azimuth and elevation angles) sampling training data from the infinite data space. The major algorithm for optimization is similar to AdaBoost~\cite{freund1997decision}, {\em i.e.}, increasing the weight of difficult samples in training the classifier.

The training process involves updating $\boldsymbol{\theta}$ and $P\!\left(\cdot\right)$ in an iterative manner. The unit that controls the classifier-sampler communication is named {\bf SampleAhead}. In each iteration, the distribution $P\!\left(\cdot\right)$ is determined by the testing results in a standalone validation set, and then used to sample a new batch of data for training the classifier (updating the parameter $\boldsymbol{\theta}$). To improve computational efficiency, we partition the entire space into a finite number of {\em buckets}. In each training epoch, the classifier is first applied on a validation set to estimate the difficulty of each bucket, and the sampler follows to construct a new training subset. This is a two-stage sampling process. Every time, a bucket is first sampled from the distribution $P\!\left(\cdot\right)$, and then a datum is sampled from the bucket following a uniform distribution. This iteration continues until the maximal number of rounds is reached.

We conduct experiments in a challenging task known as object pose estimation, which aims at predicting the viewpoint from which we capture a 2D image of an object. We use PASCAL3D+~\cite{xiang2014beyond} as the target (testing) dataset, and render a large number of synthesized images from ShapeNet~\cite{chang2015shapenet}. In comparison to the baseline approach~\cite{su2015render} which always sampled the data space from a fixed distribution, our method produces higher recognition accuracy especially in more challenging scenarios, in agreement with our motivation. In particular, when the number of extra training cases is limited, the advantage of our approach becomes even more significant.

The remainder of this paper is organized as follows. Section~\ref{sec:related_work} briefly reviews related work, and Section~\ref{sec:approach} illustrates our overall framework as well as the joint optimization approach. After experiments are shown in Section~\ref{sec:experiments}, we conclude this work in Section~\ref{sec:conclusions}.

\section{Related Work}
\label{sec:related_work}

Training machine learning systems for computer vision tasks, especially deep neural networks, often requires sufficient data to prevent over-fitting. The availability of large-scale datasets facilitates the ability of training very deep neural networks~\cite{krizhevsky2012imagenet}. However, researchers often required a considerable amount of labor to collect and annotated a large-scale dataset~\cite{deng2009imagenet,wu20143d}, or a smaller one with reasonable variability~\cite{everingham2010pascal,xiang2014beyond}.

On the other hand, the rapid development of computer graphics allows researchers to construct an unreal environment~\cite{qiu2016unrealcv, zhang2016unrealstereo}, and sample a large number of annotated synthesized data with little human labor~\cite{chang2015shapenet,johnson2017clevr}. Another possibility is to apply generative deep learning models to simulate the distribution of real data~\cite{goodfellow2014generative}. It has been verified that synthesized~\cite{su2015render,chen2016synthesizing,richardson20163d,varol2017learning} or generated~\cite{shrivastava2017learning} data are helpful in training better models. However, in either case, we are provided with an infinite space of training data, and facing the issue of making use of these synthesized data in a constrained time, {\em i.e.}, the number of sampled data is finite. A related area to this problem is named Active Vision~\cite{bajcsy1988active,blake1993active}, in which one is allowed to manipulate the viewpoint of the camera(s) in order to explore and learn richer visual knowledge from the environment. Recently, this idea was also applied to train robots in the task of visual question answering~\cite{das2017embodied,gordon2017iqa}.

There exist several ways of sampling training data from a given distribution. A straightforward solution is bootstrapping, which sampled training data with replacement. Researchers soon developed other algorithms to increase the probability of sampling a hard example, such as AdaBoost~\cite{freund1997decision} and a series of negative example mining methods~\cite{sung1996learning,rowley1998neural} to assist training in SVM~\cite{felzenszwalb2010object} and CNN~\cite{he2014spatial}. At a finer level, it is also possible to adjust the weights of different elements, so that the loss function would lean towards penalizing the errors in hard examples~\cite{simo2014fracking,loshchilov2015online,wang2015unsupervised,shrivastava2016training}. All these approaches were verified to outperform uniform sampling, especially when the easy examples occupy a considerable fraction of the data space.

In this paper, we focus on a more efficient sampling strategy. Different from previous work, we consider an infinite (continuous) data space. Instead of sampling from each instance ({\em e.g.}, an image~\cite{simo2014fracking} or a regional feature~\cite{shrivastava2016training}), we partition the entire data space into a finite number of buckets and perform two-stage sampling, detailed in Section~\ref{sec:approach:approximation}.


\section{Approach}
\label{sec:approach}

\subsection{Background and Motivation}
\label{sec:approach:background}

The goal of this work is to train an effective vision model from an infinite synthesized dataset. Throughout this paper, we assume the target model to be a classifier, denoted by $\mathbb{C}$: ${\mathbf{Y}}={\mathbf{f}\!\left(\mathbf{X};\boldsymbol{\theta}\right)}$, where $\mathbf{X}$ and $\mathbf{Y}$ are the input and output, {\em e.g.}, an image matrix and a one-hot vector, and $\boldsymbol{\theta}$ are the parameters in the model $\mathbf{f}\!\left(\cdot\right)$, {\em e.g.}, the weights of a neural network.

Training data are sampled from an image space $\mathcal{X}$. The sampling process is a function ${\mathbf{X}}={\mathbf{g}\!\left(\mathbf{U}\right)}$, where $\mathbf{U}$ are the parameters ({\em e.g.}, object position, viewpoint, lighting, {\em etc.}) required by the generator $\mathbf{g}\!\left(\cdot\right)$. Note that $\mathbf{U}$ is sampled from the parameter space $\mathcal{U}$, which is continuous and thus infinite. {\bf The core of this paper is to sample a number of $\mathbf{U}$'s at each training iteration.} Following a large corpus of previous work, we assume that each $\mathbf{U}$ is sampled independently and identically from a distribution $P\!\left(\cdot\right)$ defined in the parameter space $\mathcal{U}$. We denote the process of generating a training data by ${\mathbf{X}}={\mathbf{g}\!\left(\mathbf{U}\right)};{\mathbf{U}}\sim{P}$.

A naive example is to set $P\!\left(\cdot\right)$ to be a uniform distribution over $\mathcal{U}$, {\em i.e.}, ${P\!\left(\mathbf{U}\right)}\equiv{\mathrm{const}}$ for all ${\mathbf{U}}\in{\mathcal{U}}$. This is equivalent to generating a sufficient large synthesized dataset at the beginning, and traverse each item orderly. However, in most scenarios, the classifier are dealing with relatively easy training cases, {\em e.g.}, those cases that are already been correctly classified, so that the weights $\boldsymbol{\theta}$ cannot get trained efficiently.

\subsection{The SampleAhead Module}
\label{sec:approach:sampleahead}

\begin{figure*}[!t]
\centering
\includegraphics[width=12cm]{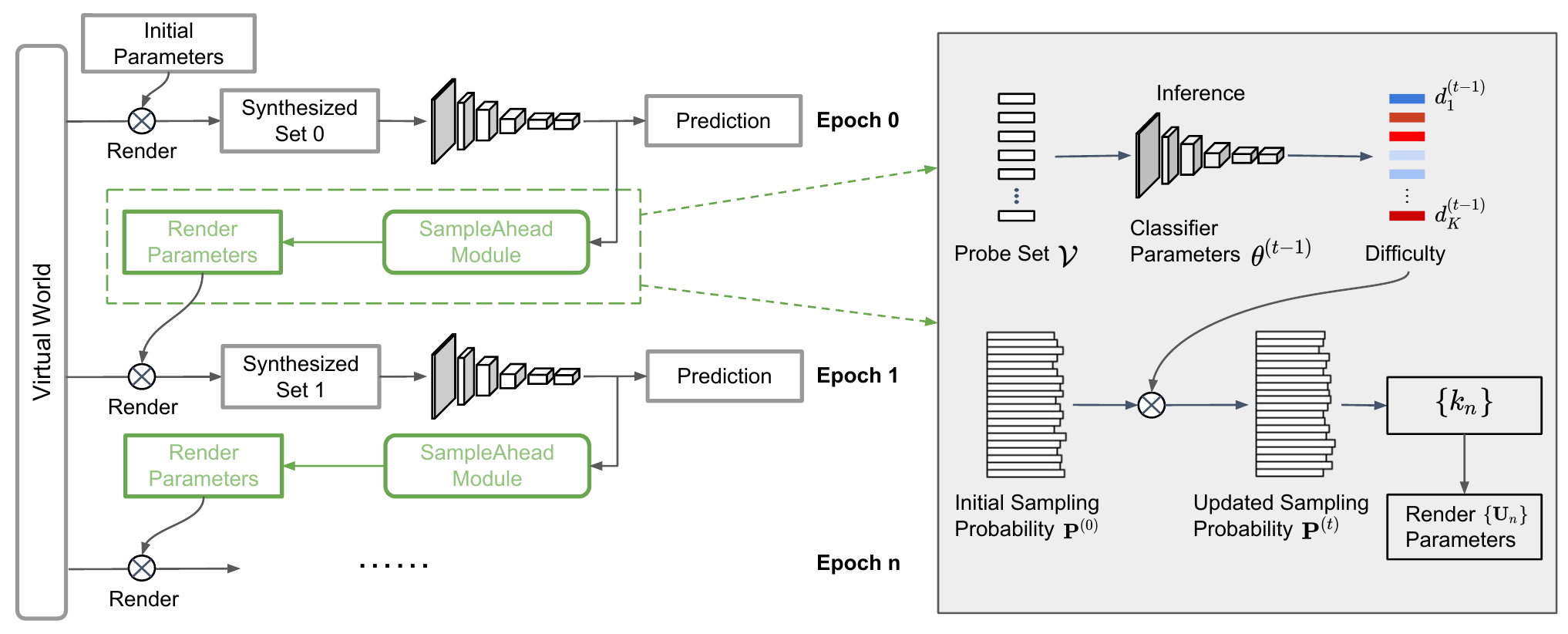}
\caption{The overall framework of our approach. Without the SampleAhead module, our approach degenerates to that fixing a synthesized dataset at the very beginning and traversing all training samples orderly. Each SampleAhead module conducts an interactive process between the classifier and the sampler, in which the classification results in a standalone probe set are used to estimate the new training data distribution, from which the sampler generates the new set for the next training epoch.}
\label{fig:sampleahead}
\end{figure*}

To deal with the above issue, we introduce a module named {\bf SampleAhead}. This module updates the data distribution $P\!\left(\cdot\right)$ {\em before} each sampling stage (in practice, before each training epoch), increasing the probability that hard examples are sampled and fed into the classifier.

Ideally, at the $t$-th iteration, for each sample $\mathbf{U}$, we hope that $P^{\left(t\right)}\!\left(\mathbf{U}\right)$ tends to have peaks at the {\em hard} cases. We start with defining the {\em difficulty} of $\mathbf{U}$, denoted by $d^{\left(t-1\right)}\!\left(\mathbf{U}\right)$, as the probability that ${\mathbf{X}}={\mathbf{g}\!\left(\mathbf{U}\right)}$ is not correctly classified by the classifier after the $t-1$-st iteration. However, directly computing $d^{\left(t-1\right)}\!\left(\mathbf{U}\right)$ for every $\mathbf{U}$ could be sensitive to noise. We make use of kernel estimation, which randomly distributes a set of $M$ probes ${\mathcal{V}}={\left\{\mathbf{V}_1,\mathbf{V}_2,\ldots,\mathbf{V}_M\right\}}$ over the entire space $\mathcal{U}$, and estimate the difficulty of $\mathbf{U}$ by:
\begin{equation}
\label{eqn:difficulty}
{d^{\left(t-1\right)}\!\left(\mathbf{U}\right)}={\frac{{\sum_{m=1}^M}d^{\left(t-1\right)}\!\left(\mathbf{V}_m\right)\cdot\omega\!\left(\mathbf{U},\mathbf{V}_m\right)}{{\sum_{m=1}^M}\omega\!\left(\mathbf{U},\mathbf{V}_m\right)}},
\end{equation}
where
\begin{equation}
\label{eqn:difficulty2}
{d^{\left(t-1\right)}\!\left(\mathbf{V}_m\right)}={1-\mathrm{Pr}\!\left[\mathbf{f}\!\left(\mathbf{g}\!\left(\mathbf{V}_m\right);\boldsymbol{\theta}^{\left(t-1\right)}\right)\mathrm{\ is\ correct}\right]},
\end{equation}
and $\omega\!\left(\mathbf{U},\mathbf{V}_m\right)$ is the weight added to $\mathbf{U}$ by the probe $\mathbf{V}_m$, {\em e.g.}, $\omega\!\left(\mathbf{U},\mathbf{V}_m\right)$ is inversely proportional to the $\ell_2$-distance between $\mathbf{U}$ and $\mathbf{V}_m$. The probe set $\mathcal{V}$ is often large in order to guarantee the coverage over the space $\mathcal{U}$.

The next step is to define the probability distribution function $P^{\left(t\right)}\!\left(\mathbf{U}\right)$ for each ${\mathbf{U}}\in{\mathcal{U}}$. Inspired by AdaBoost~\cite{freund1997decision}, we take the classification results in the previous iteration into consideration. Mathematically,
\begin{equation}
\label{eqn:probability}
{P^{\left(t\right)}\!\left(\mathbf{U}\right)}\propto{\alpha\cdot P^{\left(0\right)}\!\left(\mathbf{U}\right)+\left(1-\alpha\right)\cdot P^{\left(0\right)}\!\left(\mathbf{U}\right)\cdot e^{\beta\cdot d^{\left(t-1\right)}\!\left(\mathbf{U}\right)}},
\end{equation}
where $\alpha$ and $\beta$ are hyper-parameters. We use $P^{\left(0\right)}\!\left(\cdot\right)$ rather than $P^{\left(t-1\right)}\!\left(\cdot\right)$ to avoid the distribution from being modified too much. This strategy improves training stability.

\subsection{Approximation}
\label{sec:approach:approximation}

Note that accurately sampling from $P^{\left(t\right)}$ in Eqn~\eqref{eqn:probability} requires computing the function value at each $\mathbf{U}$, which is computationally intractable given a large $M$. Here we provide an approximation for efficient online sampling. The basic idea is to partition the entire space $\mathcal{U}$ into a finite number ($K$) of {\em buckets}, {\em i.e.}, ${\mathcal{U}}={{\bigcup_{k=1}^{K}}\mathcal{B}_k}$. Each bucket is a continuous subset of $\mathcal{U}$, and any two different buckets do not intersect with each other. Thus, we simplify Eqn~\eqref{eqn:difficulty} by only considering the probes in the same bucket, namely,
\begin{equation}
\label{eqn:approximation}
{w\!\left(\mathbf{U},\mathbf{V}_m\right)}={\mathbb{I}\!\left[\exists k;\mathbf{U}\in\mathcal{B}_k\wedge\mathbf{V}_m\in\mathcal{B}_k\right]},
\end{equation}
where $\mathbb{I}\!\left[\cdot\right]$ is the indicator function. Note that for any $\mathcal{B}_k$, every element ${\mathbf{U}}\in{\mathcal{B}_k}$ has the same distance to each probe, thus the same difficulty $d_{k}^{\left(t-1\right)}$ (omitting $\mathbf{U}$):
\begin{equation}
\label{eqn:approximated_difficulty}
{d_{k}^{\left(t-1\right)}}={\frac{{\sum_{\mathbf{V}_m\in\mathcal{B}_k}}d^{\left(t-1\right)}\!\left(\mathbf{V}_m\right)}{\left|\mathcal{V}\cup\mathcal{B}_k\right|}}.
\end{equation}
This actually leads to {\bf a two-stage sampling process}, in which a bucket-level probability is computed for each bucket:
\begin{equation}
\label{eqn:bucket_probability}
{P_k^{\left(t\right)}}={\int_{\mathbf{U}\in\mathcal{B}_k}P^{\left(t\right)}\!\left(\mathbf{U}\right)\,\mathrm{d}\mathbf{U}}.
\end{equation}
Every time we hope to generate a ${\mathbf{U}}\in{\mathcal{U}}$, we first determine the bucket index $k$ from a finite set $\left\{1,2,\ldots,K\right\}$, and then sample a $\mathbf{U}$ from $\mathcal{B}_k$ following a uniform distribution.

In practice, we update $P_k^{\left(t\right)}$ after each training epoch. This is not done after each mini-batch because of its large computational costs (requiring a complete testing in the probe set which is often large). At the beginning, $P_k^{\left(0\right)}$ is simply defined as the probability that a uniform sampling in $\mathcal{U}$ falls into $\mathcal{B}_k$. In updating $P_k^{\left(t\right)}$ with Eqn~\eqref{eqn:probability}, note that both $P^{\left(0\right)}\!\left(\mathbf{U}\right)$ and $d^{\left(t-1\right)}\!\left(\mathbf{U}\right)$ are constants within $\mathcal{B}_k$, thus Eqn~\eqref{eqn:probability} is simplified as:
\begin{equation}
\label{eqn:approximated_probability}
{P_k^{\left(t\right)}}={\alpha\cdot P_k^{\left(0\right)}+\left(1-\alpha\right)\cdot P_k^{\left(0\right)}\cdot e^{\beta\cdot d_k^{\left(t-1\right)}}}.
\end{equation}
The definition of buckets differs from case to case, and will be discussed in experiments.

Of course, there are some technical details that can be discussed, such as sharing/reusing data in the training set and the probe set, which we will investigate in the future.

\section{Experiments}
\label{sec:experiments}

\subsection{MNIST: Digit Classification}
\label{sec:experiments:mnist}

\noindent
$\bullet$\quad{\bf Dataset and Settings}

We first evaluate our approach on a toy problem, which is handwritten digit classification on the MNIST dataset~\cite{lecun1998gradient}. MNIST contains $60\rm{,}000$ training images and $10\rm{,}000$ testing images. The resolution of each image is $28\times28$. We use this relatively simple dataset to observe the behavior of our approach on a series of data augmentation as well as discover the advantage of our approach with respect to the number of training samples.

Following~\cite{cirecsan2010deep}, we consider seven types of augmentation, including digit rotation, vertical/horizontal scaling, horizontal/vertical shifting, and horizontal/vertical shearing. Each digit is processed by one and exactly one augmentation. We further partition each type into a finer stage according to the transformation parameter. The rotation angle is randomly sampled from $\left[-15^\circ,15^\circ\right]$, and it is divided into four bins $\left[-15^\circ,-7.5^\circ\right)\cup\left[-7.5^\circ,0^\circ\right)\cup\left[0^\circ,7.5^\circ\right)\cup\left[7.5^\circ,15^\circ\right]$. All other scaling/shifting/shearing parameters are divided into two bins. Thus, we obtain $16$ bins for each original training image.

The bucket set $\left\{\mathcal{B}_k\right\}_{k=1}^K$ is the Cartesian product of the class set ($10$ elements) and the bin set ($16$ elements), {\em i.e.}, there are ${10\times16}={160}$ buckets in total. This is to say, we assume, by Eqn~\eqref{eqn:approximation}, that all samples with the same class and a similar transformation share the same difficulty, which is reasonable. We randomly sample $100$ images from each bucket to compose of the probe set $\mathcal{V}$ ($16\rm{,}000$ probes in total).

\vspace{0.2cm}
\noindent
$\bullet$\quad{\bf Results and Analysis}

\newcommand{\colwidthA}{2.0cm}
\renewcommand{\tabcolsep}{0.02cm}
\begin{table*}[]
\centering
\begin{tabular}{|l||R{\colwidthA}|R{\colwidthA}||R{\colwidthA}|R{\colwidthA}|R{\colwidthA}|}
\hline
\# of Iterations & $10\rm{,}000$        & $20\rm{,}000$        & $30\rm{,}000$        & $40\rm{,}000$        & $50\rm{,}000$        \\
\hline\hline
Uniform Samp.    & $0.890\pm0.021$      & $0.835\pm0.024$      & $0.816\pm0.021$      & $0.796\pm0.018$      & $0.793\pm0.035$      \\
\hline
Our Approach     & $0.819\pm0.025$      & $0.787\pm0.019$      & $0.756\pm0.022$      & $0.758\pm0.026$      & $0.757\pm0.014$      \\
\hline
$p$-value        & $6.407\times10^{-5}$ & $2.434\times10^{-3}$ & $4.454\times10^{-4}$ & $8.097\times10^{-3}$ & $3.054\times10^{-2}$ \\
\hline
\end{tabular}
\caption{Classification error rates ($\%$) on the MNIST dataset with respect to the number of sampled images. The average and standard deviation numbers come from $10$ individual runs. The $p$-values are obtained from standard $t$-tests over these $10$ pairs.}
\label{tab:mnist}
\end{table*}

We use LeNet~\cite{lecun1998gradient} as the classifier $\mathbf{f}\!\left(\cdot;\boldsymbol{\theta}\right)$. It contains $3$ convolutional, $2$ pooling and $2$ fully-connected layers. The network is trained with Stochastic Gradient Descent. Each iteration contains a mini-batch of size $76$ ($60$ real data and $16$ synthesized data). The initial learning rate is $5\times10^{-3}$, decayed with the {\tt inv} policy, and the weight decay is fixed to be $5\times10^{-4}$. The original training process lasts for $10\rm{,}000$ iterations, but we allow a larger number ($20\rm{,}000$, $30\rm{,}000$, $40\rm{,}000$ and $50\rm{,}000$) of iterations so that more augmented data are seen. Without synthesized data, the error rate hardly goes down after $10\rm{,}000$ iterations.

Results are summarized in Table~\ref{tab:mnist}. We can observe that our approach works consistently better than the baseline approach (performing uniform sampling in the data space). In particular, when the model is constrained to see a limited amount of training data, the advantage of our approach becomes even more significant, {\em e.g.}, with $10\rm{,}000$ iterations, the absolute and relative error rate drops brought by our approach are $0.071\%$ and $7.98\%$, respectively, with a $p$-value of $6.407\times10^{-5}$, demonstrating strong statistical significance. This implies that our approach explores the data space more efficiently by aggressively looking for those challenging training samples. However, as the number of training data increases, the advantage becomes smaller, {\em e.g.}, with $50\rm{,}000$ iterations, the absolute and relative error rate drops are $0.034\%$ and $4.29\%$, respectively, with a $p$-value of $3.054\times10^{-2}$ which still suggests statistical significance. This is MNIST is relatively simple: given a sufficient amount of training data, random sampling can gradually achieve comparable performance to our approach.

\subsection{ShapeNet: Object Pose Estimation}
\label{sec:experiments:shapenet}

\noindent
$\bullet$\quad{\bf Dataset and Settings}

We move to a natural image dataset named PASCAL3D+~\cite{xiang2014beyond}, a challenging corpus for 3D object detection and pose estimation. The $12$ rigid object classes (with more than $3\rm{,}000$ images per class) in the PASCALVOC dataset~\cite{everingham2010pascal} were augmented with 3D annotations, exhibiting more variability than other 3D datasets.

Due to the limited amount of data, we follow a recent baseline named RenderForCNN~\cite{su2015render} which generated synthesized data to assist network training. To construct an augmented training set, a joint distribution of viewpoint angles and camera distances was first estimated from the PASCAL3D+ {\em real} training set, and $2.4$ million synthesized images were rendered from 3D models of ShapeNet~\cite{chang2015shapenet} following the same distribution. This is to say, the data distribution is fixed throughout the entire training process. Differently, we add the SampleAhead module (Eqn~\eqref{eqn:approximated_probability}) to enable updating data distribution according to validation.

Note that $\alpha$ in Eqn~\eqref{eqn:approximated_probability} controls the fraction of newly generated data. Setting ${\alpha}={1.0}$ causes our algorithm degenerate to the baseline, {\em i.e.}, freezing the distribution ${\mathbf{P}^{\left(t\right)}}\equiv{\mathbf{P}^{\left(0\right)}}$ throughout the entire training process. In practice, we set ${\alpha}={0.9}$ to take advantage of new data meanwhile preventing the training process from being slowed down by the time-consuming data generation (image rendering) process.

Based on these settings, we perform two challenging tasks, known as {\em object-detection-and-pose-estimation}~\cite{su2015render} and {\em viewpoint prediction}~\cite{tulsiani2015viewpoints}.

\vspace{0.2cm}
\noindent
$\bullet$\quad{\bf Object Detection and Pose Estimation}

\renewcommand{\tabcolsep}{0.09cm}
\begin{table*}
\centering
\begin{tabular}{|l|c||r|r|r|r|r|r|r|r|r|r|r||r|}
\hline
Approach            & $L$  & {\em aero.}     & {\em bicy.}     & {\em boat}      & {\em bus}       & {\em car}       & {\em chair}     & {\em table}     & {\em moto.}     & {\em sofa}      & {\em train}     & {\em tv}        & {\bf mean}      \\
\hline\hline
\cite{su2015render} & $ 4$ &          $54.0$ &          $50.5$ &          $15.1$ &          $57.1$ &          $41.8$ &          $15.7$ &          $18.6$ &          $50.8$ &          $28.4$ &          $46.1$ &          $58.2$ &          $39.7$ \\
\hline
Baseline            & $ 4$ &          $62.2$ &          $59.0$ &          $17.6$ &          $61.6$ & $\mathbf{48.2}$ &          $17.2$ & $\mathbf{20.5}$ & $\mathbf{60.0}$ & $\mathbf{34.9}$ & $\mathbf{51.6}$ & $\mathbf{60.5}$ &          $44.8$ \\
\hline
Ours                & $ 4$ & $\mathbf{63.2}$ & $\mathbf{59.8}$ & $\mathbf{18.8}$ & $\mathbf{62.7}$ &          $47.1$ & $\mathbf{19.7}$ &          $18.6$ &          $59.7$ &          $34.2$ &          $51.2$ &          $59.7$ & $\mathbf{45.0}$ \\
\hline\hline
\cite{su2015render} & $ 8$ &          $44.5$ &          $41.1$ &          $10.1$ &          $48.0$ &          $36.6$ &          $13.7$ &          $15.1$ &          $39.9$ &          $26.8$ &          $39.1$ &          $46.5$ &          $32.9$ \\
\hline
Baseline            & $ 8$ &          $55.8$ &          $52.2$ &          $14.7$ &          $48.2$ &          $42.9$ &          $15.5$ & $\mathbf{16.6}$ &          $51.0$ &          $29.3$ & $\mathbf{48.0}$ &          $45.9$ &          $38.2$ \\
\hline
Ours                & $ 8$ & $\mathbf{60.0}$ & $\mathbf{52.6}$ & $\mathbf{15.2}$ & $\mathbf{53.6}$ & $\mathbf{44.2}$ & $\mathbf{18.6}$ &          $15.3$ & $\mathbf{53.1}$ & $\mathbf{31.2}$ &          $47.4$ & $\mathbf{49.9}$ & $\mathbf{40.1}$ \\
\hline\hline
\cite{su2015render} & $16$ &          $27.5$ &          $25.8$ &          $ 6.5$ &          $45.8$ &          $29.7$ &          $ 8.5$ &          $12.0$ &          $31.4$ &          $17.7$ &          $29.7$ &          $31.4$ &          $24.2$ \\
\hline
Baseline            & $16$ &          $39.8$ &          $34.3$ & $\mathbf{ 9.4}$ &          $51.8$ &          $35.5$ &          $11.9$ & $\mathbf{17.6}$ &          $36.3$ &          $20.3$ &          $35.2$ &          $29.9$ &          $29.3$ \\
\hline
Ours                & $16$ & $\mathbf{43.6}$ & $\mathbf{40.0}$ &          $ 8.7$ & $\mathbf{57.2}$ & $\mathbf{38.6}$ & $\mathbf{14.9}$ &          $15.1$ & $\mathbf{37.5}$ & $\mathbf{23.7}$ & $\mathbf{35.6}$ & $\mathbf{36.0}$ & $\mathbf{31.9}$ \\
\hline\hline
\cite{su2015render} & $24$ &          $21.5$ &          $22.0$ &          $ 4.1$ &          $38.6$ &          $25.5$ &          $ 7.4$ &          $11.0$ &          $24.4$ &          $15.0$ &          $28.0$ &          $19.8$ &          $19.8$ \\
\hline
Baseline            & $24$ &          $28.4$ &          $23.4$ &          $ 7.7$ &          $39.5$ &          $32.1$ &          $10.6$ & $\mathbf{12.6}$ &          $28.1$ &          $19.7$ &          $38.5$ &          $17.9$ &          $23.5$ \\
\hline
Ours                & $24$ & $\mathbf{37.9}$ & $\mathbf{30.3}$ & $ \mathbf{8.7}$ & $\mathbf{47.4}$ & $\mathbf{33.9}$ & $\mathbf{13.5}$ &          $10.9$ & $\mathbf{28.7}$ & $\mathbf{22.8}$ & $\mathbf{39.0}$ & $\mathbf{26.3}$ & $\mathbf{27.2}$ \\
\hline
\end{tabular}
\caption{Accuracy ($\%$) of object detection and pose estimation. $L$ is the number of azimuth bins. A testing result is accepted if both the class and pose are predicted correctly. We use a later version released by the same authors of~\cite{su2015render} as our ``Baseline'', which performs considerably better than the original version. The {\em bottle} class is not included because its azimuth angle is almost unrecognizable.}
\label{tab:object_pose}
\end{table*}

In the first task, the system is asked to detect the object and estimate its azimuth view angle simultaneously (the elevation view angle is not considered). Following~\cite{xiang2014beyond}, the output is considered correct if it is accepted by both object detection and pose estimation. The correctness object detection is measured by the IOU between the predicted and ground-truth bounding boxes. For view angle prediction, we partition the entire $360^\circ$ azimuth range into $4$, $8$, $12$ and $24$ bins, and compute the accuracy that the predicted angle falls into the same bin as the ground-truth angle. $1$ out of of the $12$ classes ({\em bottle}) is not evaluated in this task, as the azimuth angle of such objects is unrecognizable.

The synthesized training set contain 3D objects captured from an azimuth angle of $\left[0^\circ,360^\circ\right)$ and an elevation angle of $\left[-90^\circ,90^\circ\right)$. We partition the viewpoint hemisphere into $18\times12$ bins of an equal size. Adding the $11$ classes, we have ${11\times18\times12}={2376}$ buckets in total. The validation subset from the PASCAL3D+ is used as the probe set $\mathcal{V}$.

Following the baseline~\cite{su2015render}, we extract region proposals from RCNN~\cite{girshick2014rich}, and use these images to train an AlexNet~\cite{krizhevsky2012imagenet} for joint object and viewpoint classification (${11}\times{L}$ classes, ${L}={4,8,12,24}$). All technical details (learning rate, weight decay, {\em etc.}) remain the same as the baseline. We train the network for $60\rm{,}000$ iterations, while the baseline needs $120\rm{,}000$ iterations to traverse all synthesized images. We do not update data distribution (performing uniform sampling) in the first $8\rm{,}000$ iterations so as to provide a stable initialization.

Results are summarized in Table~\ref{tab:object_pose}. In terms of average accuracy (the last column), our approach outperforms the baseline in every single task. Note that we only use half the number of iterations compared to the baseline, which demonstrates a favorable efficiency in exploring the infinite data space. Note that our baseline used on old-styled detector (RCNN) and classifier (AlexNet) which limited its accuracy, yet recent work~\cite{massa2016crafting,poirson2016fast} reported higher accuracy than our work with stronger backbones, {\em e.g.}, \cite{massa2016crafting} used Fast-RCNN for detection and VGGNet for classification. We chose to report on the same network configuration in order to make fair comparison to our baseline~\cite{su2015render}. Yet our approach is easily generalized to a wide range of network architectures.

\begin{figure}[!t]
\centering
\includegraphics[width=13cm]{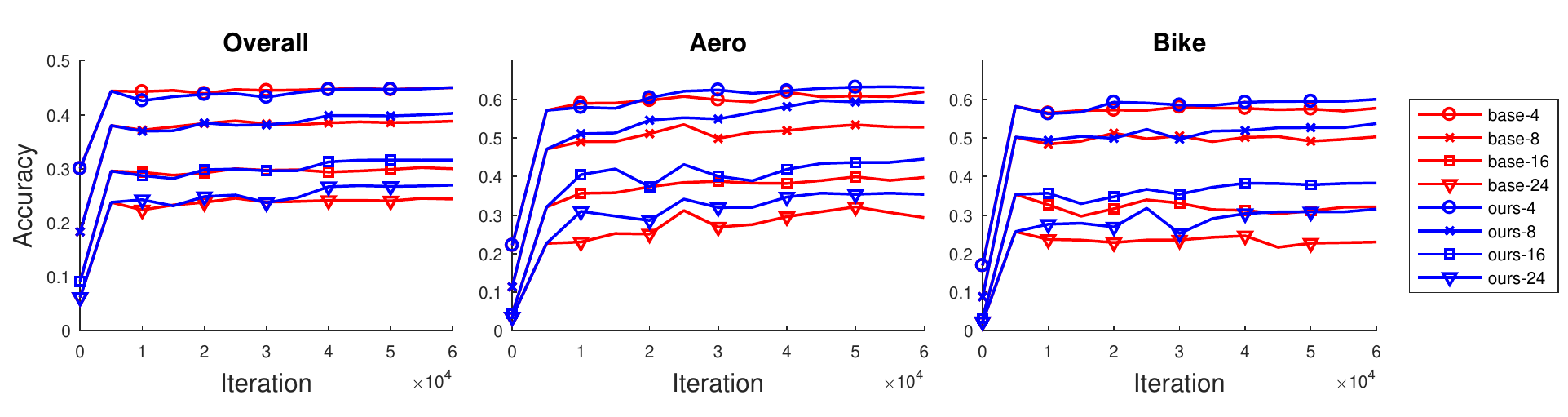}
\vspace{-0.5cm}
\caption{Compared to the baseline, our approach works better in the situation of limited data sampling. The first $8\rm{,}000$ iterations are shared by both approaches. Each number in the legend indicates the number of bins ($L$) in the classification task.}
\label{fig:curves}
\end{figure}

An interesting property of our approach is the increase in accuracy gain when the number of bins $L$ goes up. As shown in Figure~\ref{fig:curves}, this happens in both the overall accuracy and individual classes {\em e.g.}, {\em bike}. This is a side benefit brought by our approach, which mines more difficult examples to improve the performance in these challenging tasks.

\begin{figure*}[!t]
\centering
\includegraphics[width=0.24\linewidth]{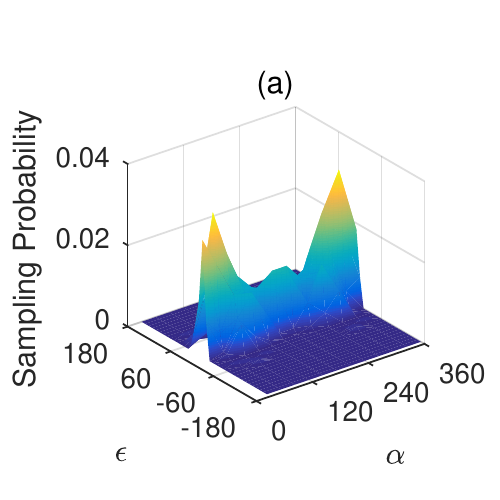}
\includegraphics[width=0.24\linewidth]{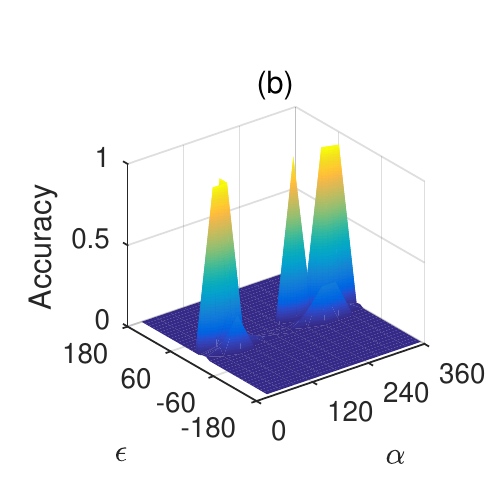}
\includegraphics[width=0.24\linewidth]{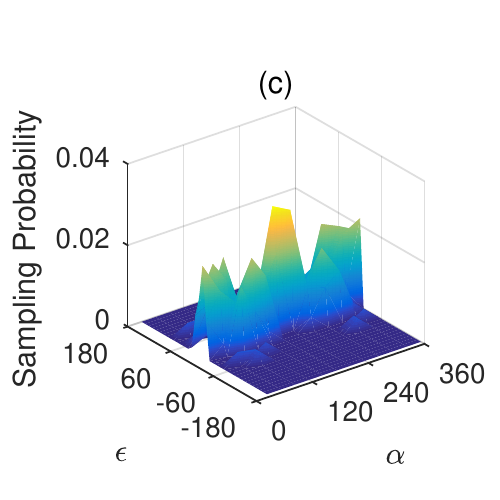}
\includegraphics[width=0.24\linewidth]{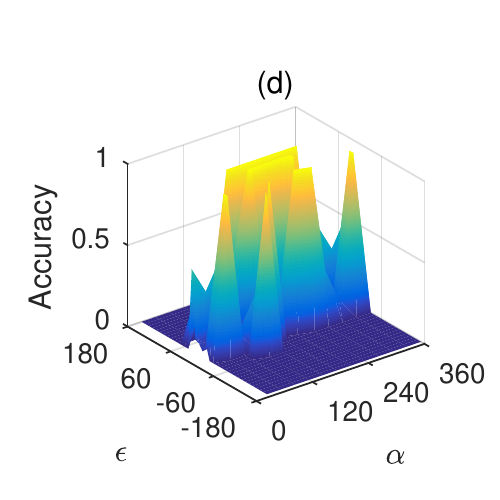}
\caption{Comparison of sampling probability and accuracy between baseline (left two) and our approach (right two). The target class is {\em bike}, and $\alpha$ and $\epsilon$ on the axes denote azimuth and elevation angles, respectively. (a) The baseline simply follows the distribution in the synthesized set, without ``realizing'' that (b) the center part (${\alpha}\approx{180^\circ}$) is more difficult. (c) On the other hand, our approach samples a larger amount of data at this area, leading to (d) a significant accuracy gain ({\em e.g.}, see Table~\ref{tab:object_pose}, $30.8\%$ vs. $23.4\%$ in $24$-bin classification).}
\label{fig:distributions}
\end{figure*}

We diagnose our approach with additional experiments. Our approach mainly benefits from two abilities, {\em i.e.}, updating sampling distribution during the training process and generating new data based on the updated distribution. Switching off the former ability turns it back to the baseline, with $0.1\%$, $0.9\%$, $1.7\%$ and $1.9\%$ accuracy drops in ${L}={4,8,12,24}$, respectively. The benefit brought by our approach becomes more significant as the number of bins goes up ({\em i.e.}, the task becomes more challenging). This is qualitatively verified in Figure~\ref{fig:distributions}, in which our approach increases the sampling probability of the difficult buckets, thus improving the overall accuracy. On the other hand, we also disable the latter ability by only allowing our approach to sample from the original $240$ million synthesized images. This causes $0.1\%$, $1.1\%$, $1.2\%$ and $1.7\%$ drops, respectively, because the synthesized dataset is fixed and the difficult class, when requiring more samples, may come into duplicated training data. This ablation study shows that both generating and sampling strategies are useful yet complementary to our approach.

As a final note, our approach does not work well on the class {\em table}, which contributes the largest deficit compared to the baseline. This class has a significant difference from others, that rotating it by $90^\circ$ merely changes its appearance, thus the $4$-bin viewpoint estimation is just a random guess. In this scenario, the baseline approach memorizes the data distribution, but our approach actually discards this ``cheating benefit'' and thus performs ``a worse guess''.

\vspace{0.2cm}
\noindent
$\bullet$\quad{\bf Viewpoint Estimation}

\renewcommand{\tabcolsep}{0.03cm}
\begin{table*}
\centering
\begin{tabular}{|l||r|r|r|r|r|r|r|r|r|r|r|r||r|}
\hline
Approach                          & {\em aero.}     & {\em bicy.}     & {\em boat}      & {\em bott.}     & {\em bus}       & {\em car}       & {\em chair}     & {\em table}     & {\em moto.}     & {\em sofa}      & {\em train}     & {\em tv}        & {\bf mean}      \\
\hline\hline
$\mathrm{Acc}_{\pi/6}$ (Baseline) &          $0.80$ & $\mathbf{0.84}$ & $\mathbf{0.62}$ & $\mathbf{0.96}$ & $\mathbf{0.95}$ &          $0.85$ &          $0.75$ & $\mathbf{0.86}$ & $\mathbf{0.88}$ & $\mathbf{0.87}$ &          $0.82$ &          $0.90$ & $\mathbf{0.84}$ \\
\hline
$\mathrm{Acc}_{\pi/6}$ (Ours)     & $\mathbf{0.84}$ & $\mathbf{0.84}$ &          $0.58$ & $\mathbf{0.96}$ &          $0.92$ & $\mathbf{0.88}$ & $\mathbf{0.91}$ &          $0.57$ & $\mathbf{0.88}$ & $\mathbf{0.87}$ & $\mathbf{0.85}$ & $\mathbf{0.93}$ & $\mathbf{0.84}$ \\
\hline\hline
$\mathrm{MedErr}$ (Baseline)      &          $10.2$ &          $12.2$ &          $18.5$ &          $ 6.5$ &          $ 4.5$ &          $ 6.4$ &          $ 12.4$ & $\mathbf{8.6}$ &          $13.0$ &          $ 11.0$ &          $ 5.7$ &          $13.1$ &          $ 10.2$ \\
\hline
$\mathrm{MedErr}$ (Ours)          & $ \mathbf{8.7}$ & $\mathbf{11.5}$ & $\mathbf{18.4}$ & $ \mathbf{6.4}$ & $ \mathbf{2.4}$ & $ \mathbf{4.5}$ & $ \mathbf{7.3}$ &          $12.5$ & $\mathbf{10.5}$ & $ \mathbf{8.3}$ & $ \mathbf{4.4}$ & $ \mathbf{9.0}$ & $ \mathbf{8.6}$ \\
\hline
\end{tabular}
\caption{Comparison on viewpoint estimation with ground-truth bounding box provided. Here, $\mathrm{Acc}_{\pi/6}$ measures accuracy (the higher the better) and $\mathrm{MedErr}$ measures error (in degrees, the lower the better). Our mean $\mathrm{Acc}_{\pi/6}$ is mainly impacted by the low value of {\em table} (explained in the texts), without which we outperform the baseline by $0.86$ vs. $0.84$.}
\label{tab:viewpoint}
\end{table*}

The second task is aimed at estimating the viewpoint to the target object. Following~\cite{tulsiani2015viewpoints}, we directly use the trained model previously, and remove the factor of inaccurate object detection by directly using the ground-truth bounding box for each object. Given the ground-truth azimuth, elevation and in-plane angles and the predicted values, we compute their rotation matrices $\mathbf{R}$ and $\mathbf{R}'$ accordingly, and the included angle between them is computed by ${\rho}={\left\|\log(\mathbf{R}^\top\mathbf{R}')\right\|_\mathrm{F}/\sqrt{2}}$, where $\left\|\cdot\right\|_\mathrm{F}$ is the Frobenius norm. There are two metrics in evaluation. The first one, named $\mathrm{Acc}_{\pi/6}$, computes the fraction that ${\rho}\leqslant{\pi/6}$; and the second one, named $\mathrm{MedErr}$, directly measures the median $\rho$ value in degrees.

Results are summarized in Table~\ref{tab:viewpoint}. Our mean $\mathrm{Acc}_{\pi/6}$ value is just slightly higher than the baseline. Note that the {\em table} class contributes negatively for the same reason analyzed in the previous task; but in all the remaining classes, our approach performs better. The average accuracies over the remaining $11$ classes are $0.86$ vs. $0.84$. In addition, the median estimation error $\mathrm{MedErr}$ is significantly reduced (a $15.7\%$ relative drop). All these experiments verify the effectiveness of our approach in learning from synthesized data.

\vspace{-0.4cm}
\section{Conclusions}
\label{sec:conclusions}
\vspace{-0.2cm}

This paper focuses on a new problem, which aims at effectively sampling synthesized data from an {\bf infinitely} large parameter space. Our motivation is very simple, {\em i.e.}, increasing the probability of generating hard examples so that the classifier gets trained better. To this end, we insert a novel module named {\bf SampleAhead}, which maintains a distribution over the sampling space. In each training unit, the distribution is first updated according to the current recognition results, and then used to sample synthesized training data and optimize the vision system. The concept of {\em buckets} is introduced to accelerate this process. Although being simple, our approach works well in a challenging vision task -- joint object detection and pose estimation, especially when the recognition task is difficult ({\em e.g.}, the number of azimuth bins is large). Our study demonstrates the effectiveness of nonuniform sampling in an infinite set, and the advantage is more significant in the scenario of less training time ({\em i.e.}, fewer synthesized data are sampled).

Our algorithm has potential applications in reinforcement learning in the real world. A typical setting is to place an agent ({\em e.g.}, a robot) in a room, and facilitate it to learn from the surrounding world by itself. Our research matches this scenario very well, since the data space is almost infinite but training time is limited.

\bibliography{egbib}
\end{document}